# Towards an AI-based knowledge assistant for goat farmers based on Retrieval-Augmented Generation

Nana Han, Dong Liu*, Tomas Norton*

Department of Biosystems, Division M3-BIORES: Measure, Model & Manage Bioresponses,

Catholic University of Leuven, Kasteelpark Arenberg 30, 3001 Heverlee, Belgium

Email: tomas.norton@kuleuven.be (T. Norton), dong.liu@kuleuven.be (Dong Liu)

Phone: +3216377531

**Abstract**

Large language models (LLMs) are increasingly being recognised as valuable knowledge communication tools in many industries. However, their application in livestock farming remains limited, being constrained by several factors not least the availability, diversity and complexity of knowledge sources. This study introduces an intelligent knowledge assistant system designed to support health management in farmed goats. Leveraging the Retrieval-Augmented Generation (RAG), two structured knowledge processing methods, table textualization and decision-tree textualization, were proposed to enhance large language models' (LLMs) understanding of heterogeneous data formats. Based on these methods, a domain-specific goat farming knowledge base was established to improve LLM's capacity for cross-scenario generalization. The knowledge base spans five key domains: Disease Prevention and Treatment, Nutrition Management, Rearing Management, Goat Milk Management, and Basic Farming Knowledge. Additionally, an online search module is integrated to enable real-time retrieval of up-to-date information. To evaluate system performance, six ablation experiments were conducted to examine the contribution of each component. The results demonstrated that heterogeneous knowledge fusion method achieved the best results, with mean accuracies of 87.90% on the validation set and 84.22% on the test set. Across the text-based, table-based, decision-tree based Q&A tasks, accuracy consistently exceeded 85%, validating the effectiveness of structured knowledge fusion within a modular design. Error analysis identified omission as the predominant error category, highlighting opportunities to further improve retrieval coverage and context integration. In conclusion, the results highlight the robustness and reliability of the proposed system for practical applications in goat farming.

# 1. Introduction

Goats production plays an important role in global agricultural systems, particularly in resource-constrained rural areas where they provide stable income, food security, and livelihood support for smallholders (Navarrete-Molina et al., 2024; Wodajo et al., 2020). According to the Food and Agriculture Organization (FAO), the global goat population exceeded 1.1 billion in 2022, including more than 200 million dairy goats. These dairy goats collectively produce an estimated 20.7 million tonnes of milk annually, representing approximately 2% of total global livestock-derived dairy production (Akshit et al., 2024; Meza-Herrera et al., 2024). Beyond their economic value, goats are increasingly recognized for their ecological adaptability and contributions to sustainable agriculture (Villarreal-Ornelas et al., 2022). On one hand, their ability to efficiently utilize marginal lands due to them being efficient and resilient grazers in arid and nutrient-poor environments (Estevez-Moreno et al., 2019). On the other hand, goat manure, which is rich in nitrogen, phosphorus, and potassium, is excreted at a rate of 0.5 to 1.2 kg per adult goat per day (Ogejo et al., 2010; Osuhor et al., 2002). Its fertilizing efficiency is comparable to synthetic fertilizers, providing a natural input for nutrient cycling and regenerative farming (Gichangi et al., 2010). This multifunctional value reinforces the strategic role of goats in developing agricultural ecosystems, and stimulates the development of Precision Livestock Farming (PLF) applications in goat farming contexts (Deepika et al., 2023).

With the ongoing intensification of livestock production, the demand for intelligent monitoring and management systems has grown markedly (Vlaicu et al., 2024). PLF has emerged as a technology-oriented paradigm aimed at enhancing animal health, welfare, and productivity by means of continuous monitoring of behaviour, physiology, and environmental parameters. This is achieved through the integration of sensor networks, imaging systems, and embedded computing technologies (Morrone et al., 2022; Norton et al., 2019). In the context of goats, PLF applications have been deployed for a range of purposes, including health surveillance (Deepika et al., 2023), behavioural analysis (Hollevoet et al., 2024), and feed intake monitoring (Chebli et al., 2022). While these systems can provide early warnings by detecting abnormal patterns or threshold violations (Gómez et al., 2021; Morrone et al., 2022), they are often limited by a lack of context-aware interpretability and seldom offer actionable, adaptive management recommendations (Islam and Scott, 2022). As a result, farmers are frequently left with unprocessed data streams or generic alarms, with little guidance on how to translate these signals into effective interventions (Kopler et al., 2023).

To bridge the data-decision gap in precision livestock farming, rule-based decision support systems (DSS) have been developed. Ethiopia's KBSGDDT (Tesfaye, 2019) and the Philippines' E-Goat Doctor (Arpay and Talirongan, 2024) are designed to provide preliminary disease diagnosis in regions where veterinary expertise is scarce, using encoded expert knowledge. However, KBSGDDT is restricted to a narrow set of goat disease profiles specific to Ethiopia's Afar region (Tesfaye, 2019), whereas E-Goat Doctor is limited to the identification of only six common goat diseases (Arpay and Talirongan, 2024). Consequently, their diagnostic performance is highly dependent on the completeness of predefined rules and the scope of the expert knowledge embedded within them. In contrast, the iSAGEDSS platform (Vouraki et al., 2020) focuses on energy and protein requirement modelling for sheep and goats. It supports European livestock producers in annual management planning by simulating hypothetical scenarios to evaluate alternative decisions. However, its deterministic structure does not account for stochastic variables. As a result, users must manually construct multiple scenarios to address unforeseen events such as extreme weather. Furthermore, iSAGEDSS lacks any integrated disease diagnosis functionality, which limits its applicability in health-related decision-making (Vouraki et al., 2020).

Recent advancements in artificial intelligence have stimulated the use of deep learning for the processing of sensor data for livestock farming applications. Convolutional Neural Networks (CNNs) have been applied for video-based postural recognition (Tung et al., 2022), body condition assessment(Temenos et al., 2024), and behavioral categorization (Gao et al., 2023). Recurrent Neural Networks (RNNs) and Long Short-Term Memory (LSTM) architectures have been applied to the temporal prediction of nutritional intake patterns (Peng et al., 2019), the detection of locomotor impairments (Bonneau et al., 2025), and the forecasting of circadian behavioural rhythms (Wagner et al., 2020). More recently, transformer-based architectures have been introduced for multivariate sequential integration in dairy cattle behaviour recognition (Zhang et al., 2025). These methods considerably outperform conventional signal-processing techniques in classification tasks. However, despite their improved perceptual capabilities, such deep learning approaches remain largely restricted to state detection or prediction. They do not, in themselves, deliver context-aware, actionable recommendations for farmers (Mahmud et al., 2021). Consequently, decision-making still depends on manual interpretation or the subsequent integration of model outputs into external decision-support frameworks (Tuyttens et al., 2022).

Large language models (LLMs), such as GPT-4, are increasingly being explored as tools for generating tailored natural-language advice from both structured and unstructured inputs (Gontijo et al., 2025; Li et al., 2025). Early agricultural applications include conversational crop advisory systems (Qing et al., 2023) and pig disease diagnosis assistants (Mairittha et al., 2025). By synthesising heterogeneous inputs, ranging from sensor data to farmer queries, these models can shift the paradigm from passive monitoring towards anticipatory decision support (Lin et al., 2024). Despite its potential, LLMs remain constrained by three persistent challenges: i) hallucination- plausible but incorrect information that can be particularly problematic in specialised contexts such as goat farming, where standardised datasets are scarce (Ji et al., 2024; Liu, 2024; Sapkota et al., 2024); ii) knowledge fragmentation -expert knowledge in this domain remains highly fragmented, dispersed across disparate formats and repositories, which further complicates reliable integration (Rudin, 2019); iii) knowledge staleness - due to their reliance on fixed training data.

Retrieval-augmented generation (RAG) (Lewis et al., 2020) offers a means of mitigating the tendency of LLMs to produce unsubstantiated outputs by grounding their responses in verified domain-specific sources. Initial livestock-focused studies report some improvements in the factual accuracy of health-related recommendations when RAG-based systems are employed (Leite et al., 2025; Menezes et al., 2024).

These frameworks retrieve relevant content from scientific literature and specialised livestock knowledge repositories, enabling the scalable generation of targeted, context-specific advice (Li et al., 2025; Samuel et al., 2025).

However, much critical knowledge in livestock farming is encoded not as conventional text but rather in structured formats such as decision trees and tabular datasets (Ekiz et al., 2020; Tajonar et al., 2022). Practitioners routinely collect and record quantitative information in tabular formats, including phenotypic traits, reproductive performance, and nutritional composition (Tajonar et al., 2022). To interpret these structured datasets, rule-based models like decision trees are widely employed, particularly for tasks such as disease diagnosis and milk yield prediction (Ekiz et al., 2020). However, traditional LLMs are poorly equipped to handle such structured data (Fang et al., 2024). Their sequential token-processing architecture is ill-suited to the order-independent nature of tables (Sui et al., 2023). When table length exceeds approximately 1,000 tokens, model performance deteriorates sharply, with attention mechanisms approaching random prediction behaviour (Fang et al., 2024). Furthermore, hierarchical relationships intrinsic to structured data are inadequately modelled, leading to weak capture of cross-column and cross-node dependencies (Liu et al., 2023). These shortcomings are particularly consequential in precision goat farming, where accurate interpretation of structured knowledge has direct implications for production efficiency, animal welfare, and farm profitability (Ekiz et al., 2020; Sintori et al., 2019). This underscores the need for specialised architectures that integrate unstructured language processing with explicit representations of structured agricultural knowledge. Such hybrid frameworks could provide the context-aware decision support required to address the complexities of real-world goat production systems.

To address these gaps, we propose a modular intelligent knowledge assistant system for goat farming. Leveraging RAG, the proposed system first developed table-to-text and decision-tree-to-text conversion methods, enabling the LLM to effectively interpret and reason over structured farming knowledge. The retrieved and converted content is then systematically organized into five core knowledge base strengthening the system's capacity to generalize across diverse decision-making scenarios. The domain-specific knowledge base including disease prevention and treatment, nutritional management, rearing management, goat milk management, and basic farming knowledge. Lastly, the real-time online retrieval module was integrated to ensure dynamic access to up-to-date industry knowledge.

Together, by integrating LLMs with the proposed dynamic knowledge retrieval RAG, this study aims to establish a robust and scalable foundation for precision decision support in goat farming.

## 2. Materials and Methods

As depicted in Fig. 1, the proposed modular architecture adopts a multi-stage, collaborative workflow, allowing flexible integration and progressive optimization. Within the context of modern goat husbandry, this design enables the incorporation of up-to-date, domain-specific knowledge - particularly valuable in the rapidly evolving context of modern goat husbandry. Specifically, the processes begins with the segmentation of domain-specific literature into semantically coherent textual units (Text Chunks). These units are then encoded into vector representations (Text Embeddings) and stored in a dedicated Vector Database. When a user submits an inquiry, it is encoded as Question Embedding using the same embedding model, allowing for similarity-based retrieval of the most relevant Text Embeddings from the Vector Database. These retrieved fragments (Top K relevant Text Chunks) are then combined with the original query to construct a structured prompt, which is subsequently input into an LLM (e.g., ChatGPT or Qwen) to generate context-aware advisory outputs.

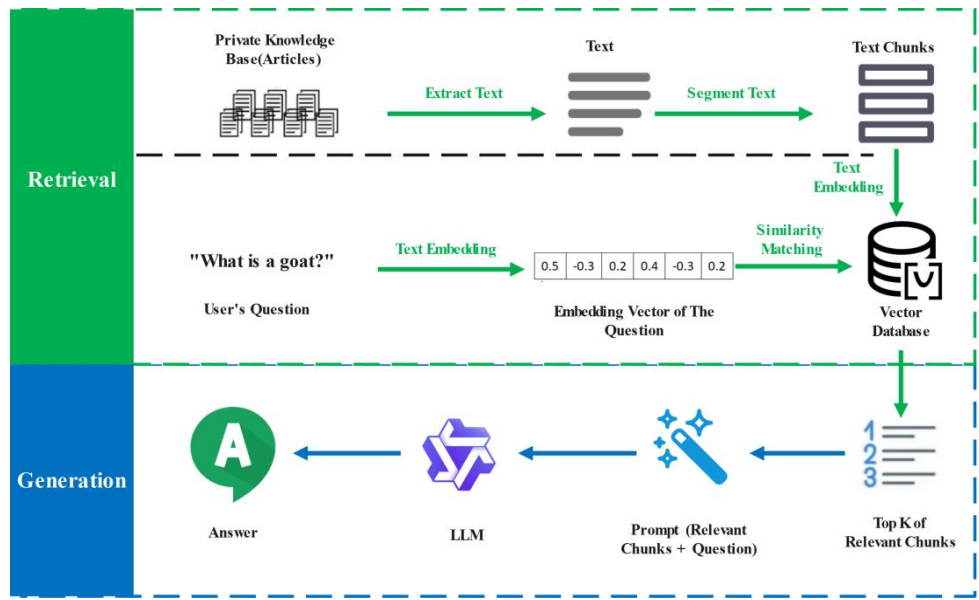

**Fig. 1.** Architecture of the RAG Framework

## 2.1 Data construction and knowledge base design

### 2.1.1 Data Sources

This study forms part of the Horizon Europe initiative Farmtopia (Farmtopia, 2023), in which a domain-specific knowledge platform - Capripedia (Govaerts & Co, 2023) is developed and used to assembling curated corpuses of 125 expert-authored articles, comprising 120 structured, text-based articles, including 56 embedded tables and five diagnostic decision trees designed by domain specialists. At present, the dataset spans a broad range of critical subdomains within goat husbandry, as illustrated in Fig. 2 - 30 articles address Disease Prevention and Treatment, 68 focus on Nutrition Management, nine cover Rearing Management, 17 pertain to Goat Milk Management, and one provides Basic Farming Knowledge. Together, these materials offer a semantically diverse and domain-relevant foundation for the development of the system's knowledge base and its retrieval-augmented question-answering component.

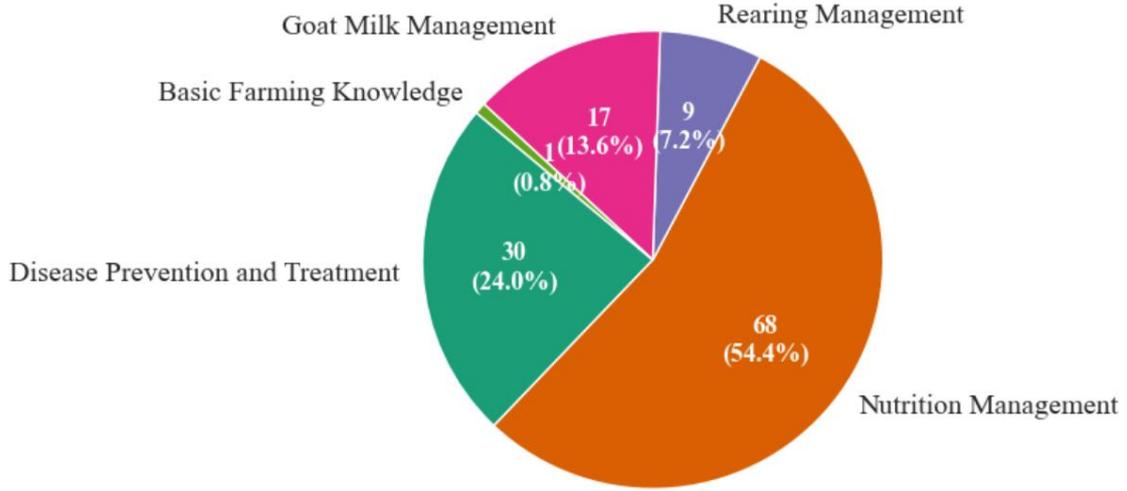

**Fig. 2.** The Distribution of Goat Farming Articles

### 2.1.2 Table textualization

Essential information, such as nutritional formulations, is systematically encoded in structured tabular formats within the goat husbandry knowledge repository. While this representation improves administrative efficiency and human readability, conventional LLMs often struggle to interpret the underlying row-column structure and classification semantics. As a result, model outputs may lack the contextual coherence required for reliable decision support.

To address this limitation, we propose a tabular textualization framework that integrates rule-based semantic parsing with a rule engine architecture. Formally, the transformation from a table $T \in R^{m \times n}$ into a textual representation can be defined as a two-stage mapping:

$$f_{\text{text}}(T) = L(R(T)) \tag{1}$$

where $R(\cdot)$ represents the rule engine responsible for preprocessing and mapping table entries, and $L(\cdot)$ denotes the LLM-driven semantic parser.

The rule engine first applies header identification, cell-level mapping, and data validation, expressed as:

$$T' = R(T) = \{(h_j, C_{ij}) \mid 1 \leq i \leq m, 1 \leq j \leq n, V(h_j, C_{ij}) = 1\} \tag{2}$$

where $h_j$ represents the column header, $C_{i,j}$ the cell content, and $V(\cdot)$ a validation function ensuring data integrity. Here, *m* denotes the total number of table rows, *n* the total number of table columns, *i* the row index, and *j* the column index.

Next, the LLM performs semantic parsing to generate context-aware textual descriptions:

$$\text{Text} = L(T') = \bigcup_{i=1}^{m} \phi\left(\{(h_j, C_{ij})\}_{j=1}^{n}\right) \quad (3)$$

where $\phi(\cdot)$ is a function that transforms each table row into a coherent natural language statement. and $\cup(\cdot)$ can denote a post-processing function for unifying terminology or resolving contextual ambiguities across rows.

The framework is designed to accommodate tables of varying dimensions and structural complexity. As shown in Fig. 3, the rule engine governs explicit mapping and validation tasks, while the LLM generates coherent narrative descriptions based on semantic context. This approach improves both the precision and interpretability of tabular-to-text conversions and provides enhanced semantic input for downstream generative tasks, while maintaining a semantic preservation constraint:

$$\forall (h_j, C_{ij}) \in T, \exists \tilde{s}_{ij} \in f_{\text{text}}(T) \text{ s.t. sem}(\tilde{s}_{ij}) \approx \text{sem}(C_{ij}) \quad (4)$$

where $\text{sem}(\cdot)$ denotes semantic equivalence.

Qwen3-235B model(Yang et al., 2025) is applied to transform 56 tables into corresponding textual representations, including 51 for nutritional management, two for disease prevention and treatment, and three for feeding management.

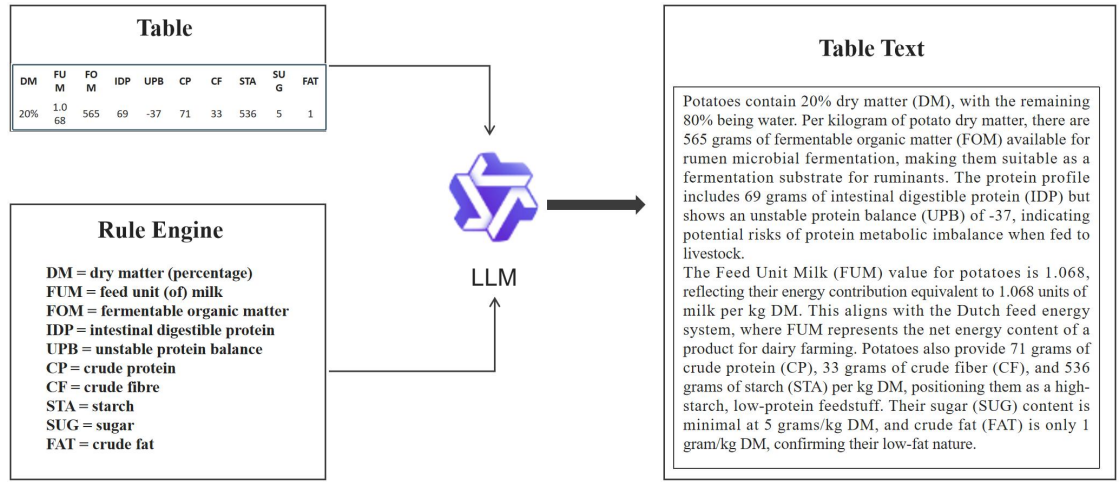

**Fig. 3.** The Process of Table Textualization

### 2.1.3 Decision-tree textualization

In conventional expert-decision system, diagnostic reasoning is typically encoded in hierarchical-tree-based formats. While well-suited for visual interpretation by humans, such structures pose significant challenges for LLMs, which struggle to interpret embedded logical relationships and conditional branches directly. As a result, model outputs may fail to capture the logical dependencies required for reliable diagnostic decision support.

To overcome this limitation, we propose a decision-tree textualisation framework that systematically converts diagnostic paths into context-rich natural language descriptions. Formally, we represent a diagnostic decision tree as a directed graph:

$$Graph = (V, E) \qquad (5)$$

where $V$ denotes the set of nodes (including internal condition nodes and terminal diagnosis nodes), and $E \subseteq V \times V$ represents the directed edges corresponding to diagnostic conditions. Each complete diagnostic path from the root node $v_{root}$ to a leaf node $v_{leaf}$ can be expressed as an ordered sequence:

$$P_k = \{v_{root}^k, v_2^k, \ldots, v_{leaf}^k\} \qquad (6)$$

Where, $k \epsilon [1, K]$, $K$ is the total number of logically complete diagnostic paths extracted from the decision tree.

The textualization process is then defined as:

$$f_{\text{text}}(G) = \cup_{k=1}^{K} \psi(P_k) \qquad (7)$$

where $\psi(\cdot)$ is a semantic transformation function that converts each path $P_k$ into a natural language description.

To support interactive diagnostic reasoning, the framework incorporates a clarification mechanism. For any user input $I \subseteq V$ that is insufficient to determine a unique diagnostic path, the system identifies all unresolved internal decision nodes and prompts the user to provide the missing information:

$$Q(I) = \{q \mid \exists v \in P_k, v \notin I\} \qquad (8)$$

where each $q$ corresponds to the missing internal decision node. A diagnosis is only returned when all required nodes in a path have been resolved:

$$\text{Diagnosis}(I) = \begin{cases} \psi(P_k), & \text{if } \forall v \in P_k, v \subseteq I, \\ Q(I), & \text{otherwise.} \end{cases} \qquad (9)$$

Finally, the interactive Q&A dataset for fine-tuning is derived as:

$$D = \bigcup_{k=1}^{K} \{(Q_{k,i}, A_{k,i})\}_{i=1}^{n_k} \tag{10}$$

where $D$ is the complete set of Q&A pairs, $Q_{k,i}$ denotes the $i$-th question on path $P_k$, $A_{k,i}$ its corresponding answer, $n_k$ the number of Q&A pairs associated with path $P_k$, $i$ the index of a Q&A pair within that path, and k the index of a unique root-to-leaf path in the decision tree.

This formulation ensures that each diagnostic path is fully captured as semantically coherent text, while preserving the original logical structure of the expert-authored decision tree, thereby improving LLM interpretability and enabling accurate, multi-turn diagnostic support. Table. 1 illustrates representative interaction patterns under different levels of user input completeness, exemplified by the subtree: Mild diarrhea → 1–3 weeks → Variable signs & lambs limping → Rota/coronavirus/Giardia.

In total, 58 logically complete subtrees were extracted from the five expert-authored decision trees. For each, interactive Q&A pairs were generated and stored as discrete textual units. This process yielded a dataset of 180 Q&A pairs, which was subsequently used for fine-tuning models and to evaluate the system's dialogue generation performance.

Table. 1. Interactive Q&A Pairs of The Subtree

| Scenario | System Response |
|---|---|
| **Complete information** | **Q**: My lamb has mild diarrhea for two weeks, shows variable signs, and is limping. <br> **A**: Consider rotavirus/coronavirus or Giardia infection. |
| **Missing "severity"** | **Q**: My lamb has intermittent diarrhea for two weeks and is limping. <br> **A**: Please tell me the severity of the diarrhea (mild or severe) so I can give more precise advice. |
| **Missing "age at onset"** | **Q**: My lamb has mild diarrhea and is limping. <br> **A**: Please tell me the duration of the illness (e.g., 1-3 weeks) to determine the possible cause. |
| **Missing "clinical pattern"** | **Q**: My lamb has diarrhea for two weeks.<br>**A**: Please describe the clinical pattern—for example, whether it's intermittent, continuous, or accompanied by limping—to choose the correct diagnostic path. |
| **Missing "severity" + "age at onset"** | **Q**: My lamb has intermittent diarrhea and is limping.<br>**A**: Please provide both the severity of the diarrhea (mild/severe) and the duration of the illness (e.g., 1–3 weeks). |
| **Missing "severity" + "clinical pattern"** | **Q**: My lamb has diarrhea for two weeks.<br>**A**: Please describe both the severity of the diarrhea (mild/severe) and any clinical signs (e.g., limping, variability). |
| **Missing "age at onset" + "clinical pattern"** | **Q**: My lamb has mild diarrhea.<br>**A**: Please tell me the duration of the illness (e.g., 1–3 weeks) and describe clinical signs (e.g., intermittent, limping). |
| **All three missing** | **Q**: My lamb has diarrhea.<br>**A**: To provide effective recommendations, please supply:1. Diarrhea severity (mild/severe);2. Duration of the illness (e.g., 1–3 weeks);3. Clinical pattern (e.g., intermittent, limping). |

### 2.1.4 Text to Q&A Pairs

Most articles exhibit a clearly defined internal structure, typically organised into chapters demarcated by capitalized subheadings. These subheadings frequently correspond to discrete knowledge units within the domain of goat farming and tend to exhibit strong internal semantic coherence. Leveraging this structural feature, we segmented each article at the level of capitalized subheadings, thereby enabling the LLM to focus on well-bounded semantic contexts during the generation of Q&A pairs. The LLM then processed each segment individually to formulate relevant questions and corresponding answers for each identified knowledge unit. These outputs were subsequently formatted into standardised Q&A pairs. The full processing pipeline is illustrated in Fig. 4.

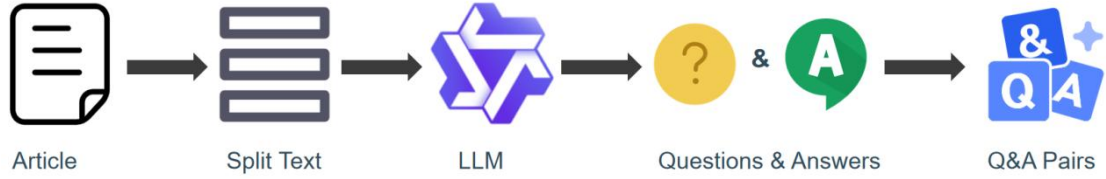

**Fig. 4.** The Process of Q&A Pair Construction

In total, 1,076 high-quality text-based Q&A pairs were automatically generated from 120 goat farming articles, along with an additional 68 Q&A pairs derived from 56 tabular data entries. This composite dataset serves as the basis for downstream model evaluation, using the Qwen-8B model (Yang et al., 2025).

### 2.1.5 Dataset Construction

The instruction fine-tuning dataset consists of three main parts: (i) standard text content from 120 goat farming articles, (ii) natural language descriptions derived from 56 structured tables, and (iii) interactive Q&A pairs from 5 expert-designed decision trees. Together, these constitute the training set.

The validation set is similarly structured into three subsets: (i) 1076 standard Q&A pairs generated from article content, used to assess general language understanding; (ii) 68 table Q&A pairs based on tabular data, aimed at evaluating the model's capacity to interpret structured information; and (iii) 180 decision tree Q&A pairs drawn from diagnostic decision trees, designed to test the model's reasoning ability in multi-step, condition-dependent scenarios.

To further assess the model's generalisation capabilities under realistic conditions, an independent test set comprising 86 representative questions was manually curated by domain experts specialising in goat health management.

The specific composition of the training set, validation set and test set is shown in Table. 2.

Table. 2. The composition of Train/Validation/Test dataset

| | | Disease Prevention | Nutrition M | Rearing | Goat Milk | Basic Farming | Total |
|---|---|---|---|---|---|---|---|
| Train | Text | 26 | 68 | 9 | 16 | 1 | 120 |
| | Tables | 2 | 51 | 3 | 0 | 0 | 56 |
|  | Trees | 4 | 0 | 0 | 1 | 0 | 5 |
|  | Total | 32 | 119 | 12 | 17 | 1 | **181** |
|  | Text Q&A | **236** | **650** | 58 | 122 | 10 | 1076 |
| Val | Table Q& | 4 | **58** | 6 | 0 | 0 | 68 |
|  | Tree Q&A | 153 | 0 | 0 | 27 | 0 | 180 |
|  | Total Q& | **393** | **708** | 64 | 149 | 10 | **1324** |
| Test | Test Q&A | 15 | 43 | 13 | 10 | 5 | **86** |

## 2.2. Dual-Retrieval Mechanism

To meet the dual needs of domain-specific expertise and up-to-date knowledge access in goat farming, a dual-retrieval mechanism was proposed as illustrated in Fig 5. It integrates: (i) a local retrieval module, which accesses a curated, domain-specific knowledge base to deliver high-relevance content; and (ii) an online retrieval module, which accesses external web resources when local content is insufficient or users explicitly request real-time updates.

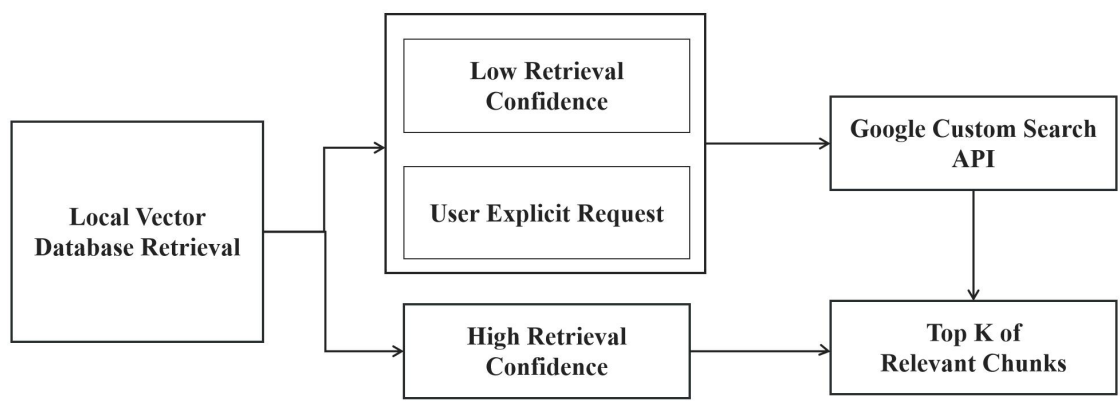

Fig. 5. The Workflow of Dual-Retrieval Mechanism

### 2.2.1 Local Vector Database Retrieval

The local knowledge base retrieval module employs vector retrieval to capture domain-specific semantic representations. A pre-trained multilingual embedding model, BGE-M3 (Chen et al., 2024), is used to encode both user queries and text chunks into fixed-length semantic vectors. BGE-M3 is selected for its strong performance in multilingual contexts, enabling accurate retrieval for users querying in different languages while maintaining semantic precision across domains.

All vectors are stored in a FAISS-based vector index (Douze et al., 2024), which supports efficient approximate nearest-neighbour search through inverted file indexing and product quantization. FAISS is chosen for its scalability and speed in high-dimensional vector spaces, allowing the system to retrieve the top-k semantically relevant text chunks based on inner product similarity. The overall workflow is illustrated in the "Retrieval" module in Fig. 1.

To balance the precision of sparse keyword-based retrieval with the broader semantic generalisation afforded by dense embeddings, the system employs a hybrid retrieval strategy that combines BM25 (Robertson and Zaragoza, 2009) scoring with cosine similarity between vector embeddings. For any user query $Q$ and candidate text chunk $D$, the comprehensive score is calculated as shown in Eq. (11):

$$\text{Score}(Q, D) = \alpha \cdot \text{BM25}(Q, D) + (1 - \alpha) \cdot \cos(\text{Vec}(Q), \text{Vec}(D)) \tag{11}$$

Here, the BM25 component is defined as follows:

$$\text{BM25}(Q, D) = \sum_{i=1}^{n} \text{IDF}(q_i) \cdot \frac{f(q_i, D)\,(k_1+1)}{f(q_i, D) + k_1\,(1 - b + b\frac{L_D}{\bar{L}})} \tag{12}$$

where $(f(q_i, D))$ is the frequency of query term $q_i$ in text chunk $D$, $L_D = |D|$ is the length of the chunk $D$ in words, $\bar{L}$ is the average chunk length in the corpus, and $k_1$, $b$ are hyperparameters typically set to 1.5 and 0.75, respectively. The inverse document frequency (IDF) is given by:

$$\text{IDF}(q_i) = \log\left(\frac{N - n(q_i) + 0.5}{n(q_i) + 0.5} + 1\right) \tag{13}$$

with $N$ being the total number of chunks, and $n(q_i)$ the number of chunks containing the term $q_i$.

$\text{BM25}(Q, D)$ captures precise keyword matching between questions and text chunks, while $\cos(\text{Vec}(Q), \text{Vec}(D))$ calculates their semantic similarity based on vector representation. By adjusting the weight parameter α (ranging from 0 to 1), the system can flexibly balance vocabulary accuracy and semantic relevance.

### 2.2.2 Online Search Module

The RAG network search module is triggered under two conditions: (i) when confidence scores from the local retrieval fall below defined thresholds (e.g., low Top-K similarity scores); or (ii) when user queries explicitly request "latest research"

or "real-time data."

It combines the original question with content retrieved from the local knowledge base to form a contextualized prompt. This combined input is then forwarded to a pre-configured Google Custom Search API (Google, n.d.), which prioritizes authoritative agricultural and veterinary sources, including open-access scientific repositories. Finally, retrieved web content is merged with the Top-K results from the local knowledge base to form a composite context, which is subsequently passed to the LLM for answer generation.

**2.3 Dynamic Augmented Prompt for Answer Generation**

This module constitutes the core of answer generation within the RAG framework. Its primary function is to dynamically integrate domain-specific knowledge of goat farming into a LLM via contextual injection, without requiring parameter fine-tuning. The overall process is illustrated in the "Generation" module of Fig. 1.

Following the knowledge retrieval process described in the preceding sections, the system constructs a structured contextual representation from the relevant text chunks. A modular prompt engineering strategy is then employed to assemble enriched prompts by combining the user query, relevant system instructions, and retrieved contextual content.

These enhanced prompts are subsequently passed to the LLM, which generates responses grounded in the injected semantic context. The entire procedure adheres to the "Retrieval–Fusion–Generation" paradigm typical of RAG-based architectures, enabling the model to incorporate external knowledge dynamically without altering its internal weights. This approach substantially improves the model's domain awareness, responsiveness, and the overall quality and reliability of generated answers in the context of goat health and farm management.

**3. Results**

**3.1 Experimental Setup**

As the present knowledge base comprises just 125 goat-farming articles, the dataset is relatively small at this stage and will be further enriched over time. To balance model performance with computational efficiency, we selected Qwen3-8B (Yang et al., 2025) as the baseline LLM. With 8 billion parameters, Qwen3-8B achieves efficient reasoning (average latency < 1 s) on a single RTX 4080 GPU while demonstrating strong generalisation capability in Q&A tasks(Touvron et al., 2023). Compared with larger models, this configuration is also better suited for fine-tuning in low-resource scenarios, mitigating the risk of overfitting while maintaining output stability and consistency (Bui et al., 2023; Seedat et al., 2023).

Based on this configuration, six comparative experiments were designed to evaluate the contributions of individual functional modules to overall Q&A performance:

- **Experiment 1:** Qwen3-8B is applied directly for question answering without retrieval or domain-specific knowledge. This serves as a knowledge-agnostic baseline for assessing the model's intrinsic adaptability to the goat-farming domain.

- **Experiment 2:** A RAG architecture is introduced. Knowledge base textual content is embedded using the BGE-M3 model, and a hybrid retrieval mechanism combining FAISS and BM25 retrieves the top three most relevant paragraphs, thereby quantifying the effect of local knowledge retrieval on Q&A accuracy, with $α = 0.3$ empirically chosen to balance semantic relevance from FAISS with lexical precision from BM25, based on prior findings and preliminary trials.

- **Experiment 3:** Building on Experiment 2, 56 structured tables are converted into natural language to assess the added value of table textualization for tasks involving structured data, such as nutritional formulation and feed management.

- **Experiment 4:** Also extending Experiment 2, interactive Q&A pairs derived from expert decision trees are incorporated. This evaluates the impact of logical chain knowledge on improving diagnostic reasoning accuracy.

- **Experiment 5:** Both table textualization and decision-tree textualization modules are integrated into the RAG framework from Experiment 2. This experiment examines the synergistic effect of multi-source structured knowledge on complex reasoning tasks.

- **Experiment 6:** Experiment 5 is further augmented with an online search module. When local retrieval confidence is low or real-time information is requested, the system dynamically activates web searches to address knowledge gaps and evaluate the effectiveness of dual-retrieval strategies.

**3.2 Evaluation Metric**

This study employs BERTScore as the primary evaluation metric. BERTScore leverages a pre-trained BERT model to generate contextualised embeddings and measures semantic similarity by computing the cosine similarity between the candidate and reference answers. In contrast to traditional n-gram-based metrics such as BLEU and ROUGE, BERTScore exhibits greater robustness to synonym substitution and syntactic variation(Zhang et al., 2019). It inherently assigns higher weights to semantically salient terms while reducing the influence of high-frequency function words on the overall score.

The precision, recall, and F1 score definitions of BERTScore are shown in Eq. (14), (15), and (16):

$$P_{BERT} = \frac{1}{|\hat{y}|}\sum_{i=1}^{|\hat{y}|} \max_{j} \cos(E(\hat{y}_i), E(y_j)) \qquad (14)$$

$$R_{BERT} = \frac{1}{|y|}\sum_{j=1}^{|y|} \max_{i} \cos(E(\hat{y}_i), E(y_j)) \qquad (15)$$

$$F1_{BERT} = \frac{P_{BERT} R_{BERT}}{P_{BERT} + R_{BERT}} \qquad (16)$$

Here, ŷ and y denote the token sequences of the model answer and the reference answer, respectively. $E(\cdot)$ represents the BERT-based embedding function, and $\cos(\cdot)$ denotes the cosine similarity function.

**3.3 Performance Comparison Across Scenarios and Datasets**

This section evaluates six instruction-fine-tuned experimental models. Each model is tested three times on both the validation and test sets, with the mean accuracy reported as the final result. BERTScore-F1 serves as the principal evaluation metric. An answer is classified as correct if its BERTScore-F1 is ⩾ 0.85; otherwise, it is deemed incorrect. This threshold was determined through manual assessment of 200 sample cases to ensure an appropriate balance between diagnostic sensitivity and false positive control.

**3.3.1 Validation Set Performance Analysis**

The validation results are shown in Table. 3 and summarized as follows:

- **Experiment 1:** As a zero-knowledge baseline, this model relies solely on the zero-shot generation capacity of Qwen3-8B. Its average accuracy on the validation set is only 9.35%, with all scenario-specific performances falling well below practical domain requirements. This outcome illustrates the model's limited adaptability to goat farming knowledge in the absence of external augmentation.

- **Experiment 2:** Introducing domain-specific intelligence via dense vector retrieval leads to a substantial accuracy improvement over Experiment 1. All scenario accuracies increase by more than 30%, yielding an overall validation set improvement of 68.09%. Notably, nutritional management shows the most striking gain, rising from 6.92% to 91.38%. These findings highlight the critical role of semantic enrichment from local knowledge repositories in mitigating LLM interpretive biases, consistent with prior research on domain adaptation in RAG-based architectures.

- **Experiment 3:** The integration of 56 structured tables provides only marginal accuracy gains (less than 1%) across scenarios. This limited impact is largely attributable to table-related queries representing only 3.09% of the validation set.

- **Experiment 4:** Mapping 58 expert decision trees into interactive Q&A pairs markedly enhances performance in the Disease Prevention and Treatment category, raising accuracy to 84.99% (+30.05%). This improvement primarily reflects the contribution of 49 decision subtrees in this domain. In Goat Milk Management, the addition of nine decision subtrees increases accuracy from 76.51% to 93.19% (+16.68%), demonstrating the value of transforming graphical diagnostic pathways into natural language conditional chains.

- **Experiment 5:** Combining table and decision tree modules further elevates overall validation accuracy to 87.73%. Performance in all scenarios exceeds baseline values by more than tenfold, confirming that multi-source structured knowledge integration effectively addresses the limitations of single-source knowledge representations.

- **Experiment 6:** The incorporation of an online search module yields negligible improvements across all scenarios. This outcome suggests that existing external resource filtering and integration methods remain suboptimal and warrant future refinement..

Table. 3. The Results of The Validation Set

| Experiment | Disease Prevention and Treatment | Nutrition Management | Rearing Management | Goat Milk Management | Basic Farming Knowledge | Average |
|---|---|---|---|---|---|---|
| Experiment 1 | 6.36 | 6.92 | 6.07 | 7.38 | 20.00 | 9.35 |
| Experiment 2 | 53.68 | 91.38 | 85.65 | 76.51 | 80.00 | 77.44 |
| Experiment 3 | 53.94 | **91.81** | 85.93 | 78.52 | 80.00 | 78.04 |
| Experiment 4 | **84.99** | 91.10 | 83.75 | 93.19 | 80.00 | 86.61 |
| Experiment 5 | 84.73 | 91.53 | 89.09 | 93.29 | 80.00 | 87.73 |
| Experiment 6 | 84.68 | 91.85 | **89.21** | **93.75** | 80.00 | **87.90** |

### 3.3.2 Generalization Performance on the Test Set

The test set results are shown in Table 4. Accuracy in "Disease Prevention and Treatment" and "Nutritional Management" closely mirrors that of the validation set, indicating that the modular system design enables strong generalization of domain knowledge to previously unseen data. For "Rearing Management", test accuracy reached 100%, which can be attributed to the comprehensive coverage of core knowledge in the underlying knowledge base. However, in the "Goat Milk Management", the best test accuracy was only 60%, which is much lower than the validation set. This shows that some test questions involve new or emerging topics that are not covered in our articles, and for dual-retrieval, we need better external source filtering strategies. The knowledge base lacks some basic information, but after adding online search, the accuracy is improved to 80%. Overall, the model achieved an average accuracy of 84.22% on the test set, demonstrating the ability to generalize to new and previously unseen questions.

These findings reinforce the effectiveness of the RAG framework in integrating domain-specific knowledge for goat farming. The table and decision tree textualization modules further enhance the model's capacity to interpret structured data and perform multi-conditional logical reasoning. When applied jointly, this heterogeneous knowledge integration strategy yields the most consistent and accurate performance across all experimental scenarios.

Table. 4. The Results of The Test Set

| Experiment | Disease Prevention and Treatment | Nutrition Management | Rearing Management | Goat Milk Management | Basic Farming Knowledge | Average |
|---|---|---|---|---|---|---|
| Experiment 1 | 20.00 | 4.65 | 0.00 | 10.00 | 20.00 | 10.93 |
| Experiment 2 | 80.00 | 86.05 | 92.31 | 30.00 | 60.00 | 69.67 |
| Experiment 3 | 86.67 | 79.07 | **100**.00 | 40.00 | 60.00 | 73.15 |
| Experiment 4 | 83.33 | 86.05 | 92.61 | 40.00 | 60.00 | 72.40 |
| Experiment 5 | 86.67 | 87.60 | **100**.00 | 50.00 | 60.00 | 77.47 |
| Experiment 6 | **91.19** | **89.92** | **100**.00 | **60**.00 | **80**.00 | **84.22** |

### 3.4 Impact of Modular Design on Different Types of Q&A Tasks

To systematically evaluate the contribution of each module to the intelligent goat-farming knowledge assistant system, four Q&A task types were defined: Text Q&A, Table Q&A, Tree Q&A, and New Q&A. The first three task types were derived from the validation set, while New Q&A was based on ten previously unseen questions from the test set that are not covered by the knowledge base. The results are illustrated in Fig. 6.

**Text Q&A:** Experiment 1 achieved only 10.55% accuracy, highlighting the baseline model's limited capacity to interpret goat farming knowledge. After incorporating local knowledge retrieval in Experiment 2, accuracy surged to 88.70% (+78.15%), confirming the central role of domain-specific retrieval and aligning with prior evidence of RAG's effectiveness in domain adaptation. Experiments 3–6 maintained approximately 88% accuracy, indicating that retrieval alone is sufficient for text-based tasks, while additional structured modules yield only marginal gains.

**Table Q&A:** Experiment 2 improved accuracy from 2.30% to 58.62%, demonstrating that local retrieval partially supports tabular knowledge utilisation. However, implementing table textualization in Experiment 3 further increased accuracy to 88.79% (+30.17%), confirming its decisive role in enabling LLMs to process structured tabular information. In contrast, Experiment 4 showed a drop to 56.04%, indicating that decision tree knowledge offers little value for table-related reasoning. Experiments 5–6 subsequently restored accuracy to approximately 86% through multi-source knowledge integration, while online search (Experiment 6) provided negligible incremental benefit.

**Tree Q&A:** Experiments 1-3 produced near-zero accuracy, demonstrating that conventional retrieval mechanisms are insufficient for complex multi-conditional reasoning. In Experiment 4, converting 58 expert decision trees into interactive Q&A pairs increased accuracy dramatically to 87.03%, underscoring the necessity of decision tree textualization for diagnostic reasoning tasks. Experiments 5-6 retained comparable performance, with online search again offering minimal additional value in this context.

**New Q&A:** Experiments 2-5 maintained 50% accuracy, suggesting that answering emerging or knowledge-base-absent queries requires real-time external information. Experiment 6, which activated online search, raised accuracy to 60%, demonstrating its potential to address knowledge gaps. Nevertheless, these results also point to the need for improved strategies for filtering and prioritising external information sources to ensure reliability.

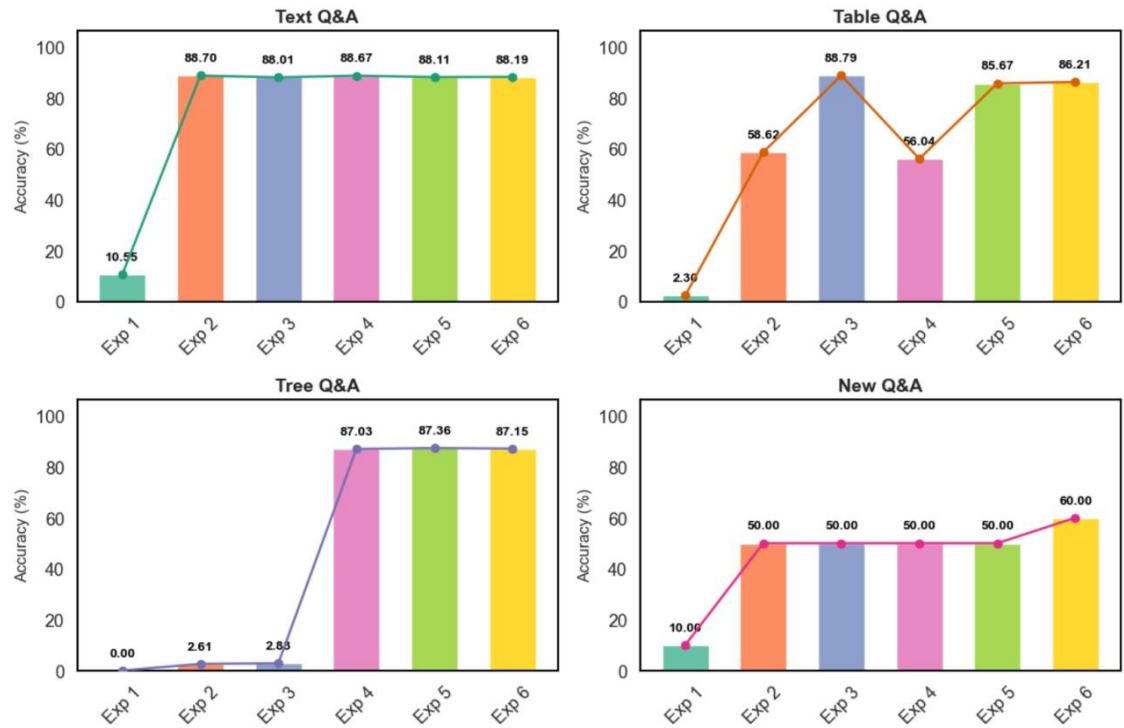

Fig. 6 The Results on Different Types of Q&A Tasks

**3.5 Analysis of Error Types in the Test Set**

To better understand the system's performance variability and its potential limitations across different task scenarios, we conducted an error classification and statistical analysis based on repeated errors observed in three independent test runs of Experiment 5. Errors were categorised into four types (Magesh et al., 2025; McKenna et al., 2023):

- **Omission:** Key knowledge points are absent from the response, leading to incomplete or insufficient answers.

- **Hallucination:** The generated content contains factual inaccuracies or unsupported information.

- **Unsupported reasoning:** The model provides logically plausible answers that, due to insufficient context, fail to address the core of the question.

- **Non-error:** Semantically correct responses misclassified as errors when their BERTScore-F1 marginally falls below the 85% threshold. These cases typically arise from lexical or phrasing differences rather than genuine content inaccuracies.

The distribution of error types across the five application scenarios is presented in Fig.

7.

Omission emerged as the predominant error category, accounting for 8 cases (50%), primarily in Disease Prevention and Treatment, Nutrition Management, and Basic Farming Knowledge. This pattern suggests gaps in retrieval coverage and contextual integration. Addressing this issue will likely require optimising vector retrieval strategies, fine-tuning the Top- K threshold, and refining prompt templates to ensure more comprehensive inclusion of relevant knowledge.

Hallucination was observed in 3 cases (18.75%), all within Goat Milk Management. These errors stemmed from the inclusion of low-quality or irrelevant information retrieved via web search. To mitigate this, future improvements should include robust filtering of external sources, the integration of knowledge graphs, and systematic credibility assessment mechanisms to reduce noise.

Unsupported reasoning accounted for 2 cases (12.5%), typically arising from fragmented retrieval or weak semantic links within the knowledge base. These errors indicate the need for restructuring knowledge storage to enhance semantic cohesion and revising context aggregation strategies to generate more coherent, well-supported reasoning chains.

Non-errors (3 cases, 18.75%) fell within an acceptable range and had negligible practical impact. Notably, the Rearing Management scenario exhibited zero errors, reflecting strong system performance in this domain, likely due to complete knowledge base coverage and relatively straightforward task logic.

In summary, this error analysis highlights three key areas for system optimisation: (i) expanding knowledge coverage, (ii) improving contextual integration during retrieval, and (iii) refining external information filtering. These findings provide a clear roadmap for enhancing both the accuracy and reliability of the intelligent knowledge assistant system.

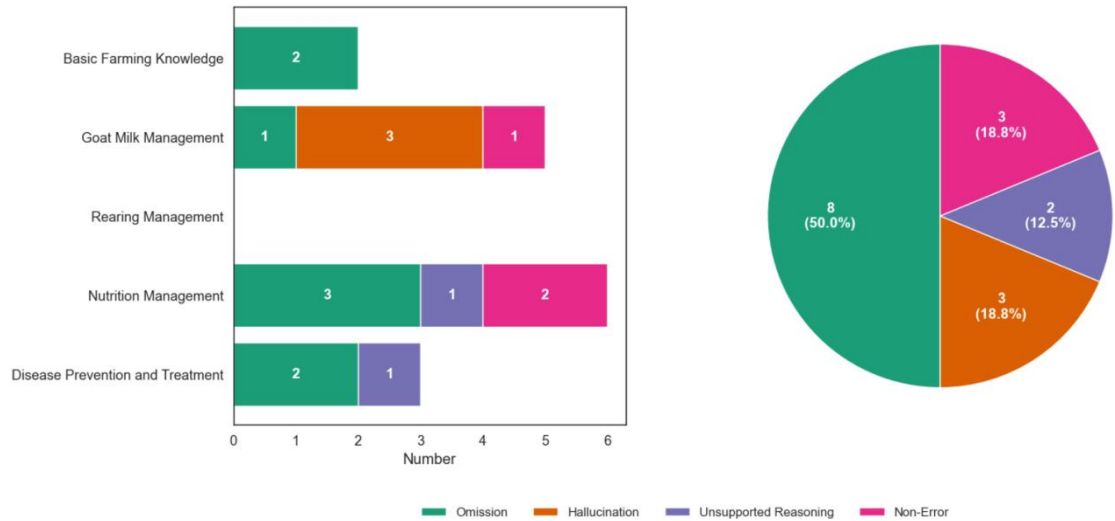

Fig. 7 The Results on Different Types of Q&A Tasks

## 4. DISCUSSION

This study presents a modular, RAG-based intelligent knowledge assistant system for goat farming, designed to integrate multi-source knowledge for enhanced contextual reasoning. The system was comprehensively evaluated using both a structured validation set and an independently curated test set. Evaluation results demonstrate that each core module, namely, local knowledge retrieval, tabular data textualization, and decision tree transformation, offers distinct functional strengths aligned with specific task demands. When deployed in combination, these modules exhibit synergistic effects, yielding superior overall performance across all evaluated task categories.

### 4.1 Technical Effectiveness Analysis

The local knowledge base retrieval module demonstrated a substantial improvement over the baseline, with an overall accuracy increase of nearly 70% on the validation set. This result confirms that, in the absence of domain-specific prompting, LLMs exhibit notable knowledge deficiencies in goat farming tasks. Vector retrieval proved effective in extracting relevant information dispersed across articles, thereby supplying accurate and contextually appropriate input for the generation module. When integrated with other modules, the system's heterogeneous knowledge architecture achieved a balanced performance across task types, attaining a validation accuracy of 87%.

The table textualisation module improved Q&A accuracy on table-based tasks by approximately 30% by converting structured tabular data into natural language format. However, due to the relatively small number of tables in the validation set, its overall contribution to system-wide performance remained modest. By contrast, the decision tree textualization module significantly enhanced the system's multi-conditional reasoning capabilities, leading to marked improvements in categories such as Disease Prevention and Treatment and Goat Milk Management. Specifically, accuracy on decision tree-based Q&A tasks increased from 0% to 87.03%, underscoring the value of tailored textualization strategies for structured knowledge formats.

The dual-retrieval module, which dynamically integrates external web content, further improved accuracy in answering novel or emerging questions, reaching 60% on unseen queries. This result affirms the importance of supplementing static internal knowledge with real-time external information in contexts where the knowledge base may be incomplete or outdated.

Error analysis on the test set revealed stable system behaviour in Rearing Management tasks, without obvious errors. In other domains, most errors were attributable to minor omissions or occasional hallucinations, while errors due to unsupported reasoning remained relatively infrequent. The proportion of non-error cases was within acceptable bounds. Taken together, these findings highlight the system's robustness and potential for practical deployment in real-world goat farming scenarios.

**4.2 Limitations and Suggestions for Improvement**

While the online search module mitigates some gaps in the static knowledge base, the quality and reliability of externally retrieved content remain variable. To enhance both the credibility and temporal relevance of retrieved information, future work may explore the integration of structured knowledge graphs or the development of dynamic filtering mechanisms based on source reliability scoring.

Performance in the Goat Milk Management domain remains somewhat inconsistent. This may be addressed by expanding the training dataset through active learning approaches and refining the system's dynamic weighting strategy for integrating heterogeneous knowledge sources, thereby improving model stability in domain-specific tasks.

Finally, current system evaluation has been limited to text-based data. Future research will explore multimodal integration, incorporating visual and auditory inputs, to support a more comprehensive, context-aware decision-support framework applicable to real-world goat farming operations.

## 5. Conclusions

Building on the RAG framework, this study designed and validated an intelligent knowledge assistant system for goat farming, introducing two complementary methods for structured knowledge processing: table textualization and decision-tree textualization. The decision-tree textualization method converts expert decision trees into natural language conditional chains, transforming complex multi-conditional reasoning into interactive Q&A pairs. This approach enhances the model's capacity for logical inference in tasks such as disease diagnosis.

Using six comparative experimental paradigms, the system evaluated the individual and combined contributions of local knowledge retrieval, table textualization, decision-tree textualization, and dual-retrieval mechanisms across multiple task types. The results were clear: the system achieved a peak average accuracy of 87.90% on the validation set and 84.22% on the test set, demonstrating strong generalisation capability. For text, table, and tree Q&A tasks, accuracy consistently exceeded 85%, confirming the value of integrating local knowledge, structured knowledge textualisation, and modular retrieval. For novel Q&A tasks, dual-retrieval achieved 60% accuracy, suggesting that while online retrieval shows potential for addressing knowledge gaps, further refinement in information filtering and source credibility assessment remains essential.

Error analysis of the test set indicated that omission errors were the most prevalent category, accounting for approximately 50% of all errors. This highlights the need to improve retrieval recall and strengthen contextual integration within the system. Other error types, hallucination, unsupported reasoning, and non-error, collectively represented less than 20% of the total, keeping the overall error distribution within an acceptable range. Notably, a subset of "non-error" cases was misclassified during automated evaluation due to threshold-related bias; however, manual verification confirmed their correctness, reinforcing the system's reliability for practical use.

Future work will prioritize the development of a multimodal intelligent system capable of processing text, audio, and image inputs. In parallel, the integration of knowledge graphs will be pursued to improve the accuracy of external content retrieval, while active learning strategies will be employed to continuously expand the domain-specific dataset. Together, these advancements aim to enhance the system's practicality, robustness, and adaptive intelligence throughout the goat farming workflow.

**CRediT authorship contribution statement**

Nana Han: Writing – review & editing, Writing – original draft, Visualization, Validation, Software, Methodology, Investigation, Formal analysis, Data curation, Conceptualization. Dong Liu: Writing – review & editing, Methodology, Formal analysis. Tomas Norton: Writing – review & editing, Supervision, Resources, Project administration, Methodology, Funding acquisition, Formal analysis, Conceptualization.

**Declaration of competing interest**

The authors declare that they have no known competing financial interests or personal relationships that could have appeared to influence the work reported in this paper.

**Funding**

This work was supported by the European Horizon project (Farmtopia, Grant agreement lD: 101083541), funded under the call HORIZON-CL6-2022-FARM2FORK-02-two-stage.

**Acknowledgements**

The authors would like to thank Govaerts & Co. for providing the goat farming article. The authors would also like to thank Wim Govaerts, Lander Govaerts, and Willem Govaerts for providing the test set questions. We also acknowledge the financial support of the European Union for KUL.

**Data availability**